\documentclass[a4paper,fleqn]{cas-dc}
\usepackage[numbers]{natbib}
\usepackage{algorithmic}
\usepackage{algorithm}
\usepackage{colortbl}
\usepackage{booktabs}
\usepackage{amsmath,amsfonts}
\usepackage{listings}
\usepackage{multirow}
\usepackage{array}
\usepackage[caption=false,font=normalsize,labelfont=sf,textfont=sf]{subfig}
\usepackage{textcomp}
\usepackage{stfloats}
\usepackage{url}
\usepackage{verbatim}
\usepackage{graphicx}
\definecolor{mygray}{rgb}{0.92,0.92,0.92}

\def\tsc#1{\csdef{#1}{\textsc{\lowercase{#1}}\xspace}}
\tsc{WGM}
\tsc{QE}
\tsc{EP}
\tsc{PMS}
\tsc{BEC}
\tsc{DE}

\begin{document}
\let\WriteBookmarks\relax
\def\floatpagepagefraction{1}
\def\textpagefraction{.001}
\shorttitle{FViT: A Focal Vision Transformer with Gabor Filter}

\shortauthors{Yulong Shi et~al.}

\title [mode = title]{FViT: A Focal Vision Transformer with Gabor Filter}                      

\author [1]{Yulong Shi}[style=chinese]

\author[1]{Mingwei Sun}[style=chinese]
\cormark[1]
\ead{smw_sunmingwei@163.com}

\author[2]{Yongshuai Wang}[style=chinese]

\author[1,3]{Zengqiang Chen}[style=chinese]

\affiliation[1]{organization={College of Artificial Intelligence},
	addressline={Nankai University}, 
	postcode={300350}, 
	state={Tianjin},
	country={China}}
	
\affiliation[2]{organization={School of Artificial Intelligence},
	addressline={Tiangong University}, 
	postcode={300387}, 
	state={Tianjin},
	country={China}}

\affiliation[3]{organization={The Key Laboratory of Intelligent Robotics of Tianjin},
                addressline={Nankai University}, 
                postcode={300350}, 
                city={Tianjin},
                country={China}}

\cortext[cor1]{Corresponding author}

\begin{abstract}
Vision transformers have achieved encouraging progress in various computer vision tasks. A common belief is that this is attributed to the capability of self-attention in modeling the global dependencies among feature tokens. However, self-attention still faces several challenges in dense prediction tasks, including high computational complexity and absence of desirable inductive bias. To alleviate these issues, the potential advantages of combining vision transformers with Gabor filters are revisited, and a learnable Gabor filter (LGF) using convolution is proposed. The LGF does not rely on self-attention, and it is used to simulate the response of fundamental cells in the biological visual system to the input images. This encourages vision transformers to focus on discriminative feature representations of targets across different scales and orientations. In addition, a Bionic Focal Vision (BFV) block is designed based on the LGF. This block draws inspiration from neuroscience and introduces a Dual-Path Feed Forward Network (DPFFN) to emulate the parallel and cascaded information processing scheme of the biological visual cortex. Furthermore, a unified and efficient family of pyramid backbone networks called Focal Vision Transformers (FViTs) is developed by stacking BFV blocks. Experimental results indicate that FViTs demonstrate superior performance in various vision tasks. In terms of computational efficiency and scalability, FViTs show significant advantages compared with other counterparts.
\end{abstract}


\maketitle

\section{Introduction}
In recent years, vision transformers have made impressive progress in various computer vision tasks such as image classification~\cite{zhang2023single, goceri2024vision}, object detection~\cite{zhang2024vision, jiang2024vision}, and semantic segmentation~\cite{zheng2023sketch, xu2024mctformer+, islam2024comprehensive}. Vision transformers provide new paradigms and solutions for these tasks, thereby challenging the dominance of Convolutional Neural Networks (CNNs) in the field of computer vision~\cite{papa2024survey, yar2024modified}. The prevailing view is that the key factor for the success of vision transformers is self-attention~\cite{rao2023gfnet}. This attention mechanism can simultaneously model the relationships between different feature tokens in the input sequence, showcasing excellent performance in global feature interaction and long-range dependency extraction~\cite{li2024scformer, zhang2024catnet}.

Nonetheless, self-attention continues to face several inherent challenges and limitations. (1) Compared with convolution, self-attention has quadratic computational complexity and high memory cost, which is especially noticeable when dealing with high-resolution images and videos. (2) Self-attention tends to focus on the overall information of input sequences and lacks sensitivity in processing local features and details of targets, thereby affecting its performance in dense prediction tasks. (3) Self-attention lacks desirable induction bias, which means it requires large amounts of training data for optimization. Especially in scenarios with limited data, the risk of overfitting in vision transformers is increased. To alleviate these issues, various self-attention variants have been proposed~\cite{liu2023efficientvit, liu2024convolution, huang2023vision, zhou2024ristra, zhu2023biformer, chen2024hint, liu2024degradation, koyuncu2024efficient}, FAT~\cite{fan2024lightweight} and BiFormer~\cite{zhu2023biformer} are two representatives. They introduce modulated convolution and sparse adaptive queries to reduce the computational cost of self-attention, respectively. Another impressive work is CrossFormer++~\cite{wang2023crossformer++}, which designs a Long-Short Distance Attention (LSDA) for vision transformers. This LSDA splits the self-attention module into a short-distance attention and a long-distance attention, thereby reducing the computational burden while preserving the rich small-scale and large-scale features of the targets.

\begin{figure*}[t]
	\centering         
	\includegraphics[width=\textwidth]{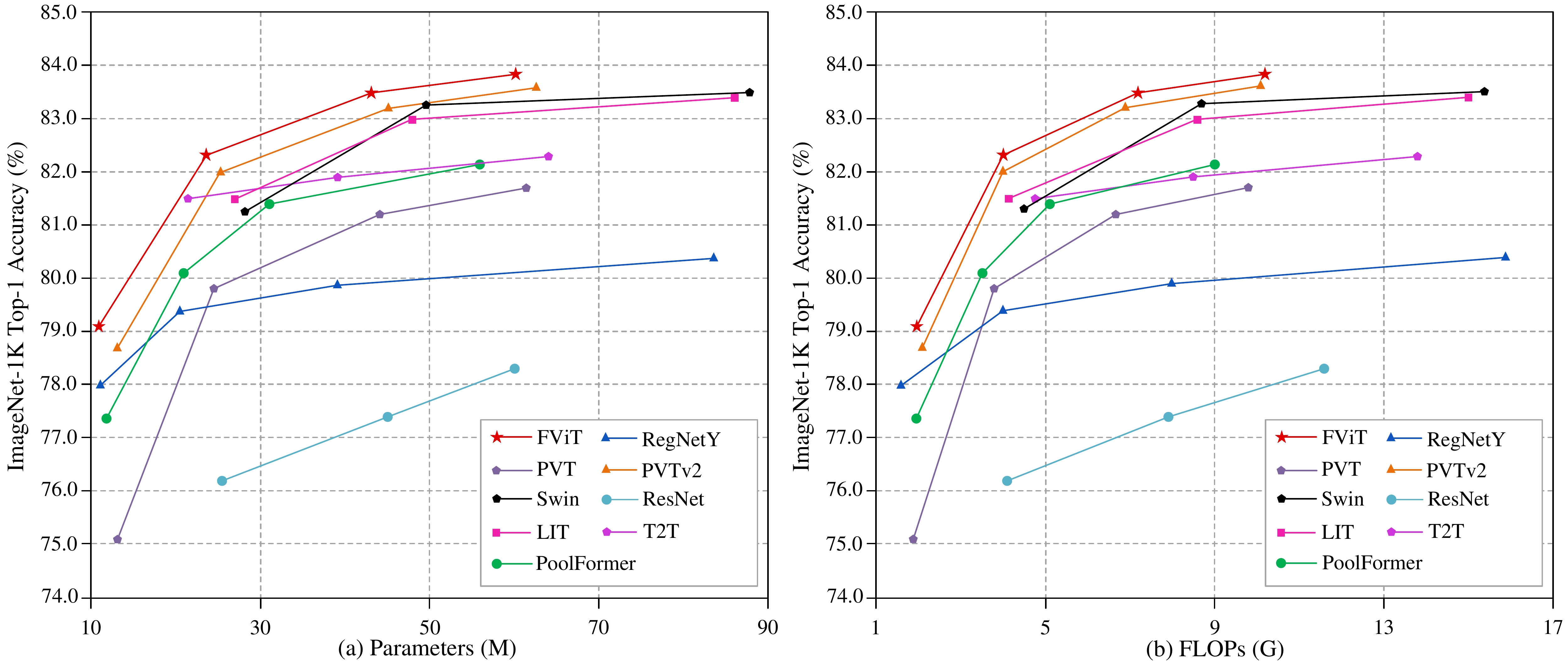}
	\caption{Comparison of Top-1 accuracies achieved by the FViTs and other baselines on the ImageNet-1K dataset. Compared with other counterparts, FViTs achieve a better trade-off in terms of parameters, computational complexity and performance on the ImageNet-1K classification task.}
	\label{fig:flops_pamer}
\end{figure*} 

Although the above mentioned methods such as downsampling and sparsification can alleviate the issues of high computational complexity and memory cost in self-attention. However, they also lead to the loss of important information, thereby resulting in incomplete feature representations. Up to now, self-attention is still one of the most effective techniques in deep learning. With limited computing resources, a question must be raised. Do we really need self-attention, and do we have better alternatives? Several recent works~\cite{lee2021fnet, rao2023gfnet, yao2022wave, yu2022metaformer, yan2024diffusion, touvron2022resmlp, garg2024gestformer} have given their respective answers. FNet~\cite{lee2021fnet} replaces self-attention with the standard non-parametric Fourier transform, proposing a fast and efficient transformer model. It achieves 92\% and 97\% of the accuracy of BERT-Base and BERT-Large on the GLUE benchmark, respectively. MetaFormer~\cite{yu2022metaformer} is another influential work that studies vision transformers from a general architecture perspective. It demonstrates that the success of vision transformers is not solely attributed to self-attention, but also benefits from the structure of vision transformers. Other works~\cite{pei2024efficientvmamba, shi2024multi, liu2024vmamba}, such as Vision Mamba~\cite{liu2024vmamba}, capture the long-range dependencies between feature tokens by adopting state-space models, achieving linear computational complexity and showing broad application prospects in computer vision tasks. Their findings provide inspiration for our research. 

In this paper, the potential advantages of combining vision transformers with classical image processing algorithms are revisited, and an effective Learnable Gabor Filter (LGF) is designed using convolution. As an alternative to self-attention, the LGF exhibits high efficiency and scalability. In addition, drawing inspiration from neuroscience, a Dual-Path Feed Forward Network (DPFFN) is introduced to emulate the parallel and cascaded information processing scheme of the biological visual cortex. Based on LGF and DPFFN, a Bionic Focal Vision (BFV) block is proposed as the basic building block. The structure of BFV is found to be strikingly similar to the computer simulation model of visual attention~\cite{itti1998model} first proposed in the 1990s. The only difference is that there is no carefully designed winner-take-all competition mechanism at the end of the BFV block. Furthermore, following the hierarchical design concept~\cite{li2023uniformer, shi2023evit, wang2022pvt}, a unified and efficient family of pyramid backbone networks called Focal Vision Transformers (FViTs) is developed by stacking BFV blocks. The FViTs comprises four variants, namely, FViT-Tiny, FViT-Small, FViT-Base and FViT-Large. These variants are designed to enhance the applicability of FViTs for various computer vision tasks. Fig.~\ref{fig:flops_pamer} shows the performance comparison of FViTs with other self-attention and non-self-attention baselines on the ImageNet-1K~\cite{russakovsky2015imagenet} dataset. Compared with other counterparts, FViTs achieve a better trade-off between parameters, computational complexity and performance on the ImageNet-1K classification task.

The main contributions are summarized as follows.

(1) As an alternative to self-attention, an efficient Learnable Gabor Filter (LGF) based on convolution is proposed. It is employed to simulate the response of fundamental cells in the biological visual system to input images, thereby prompting models to focus on discriminative feature representations of targets from various scales and orientations.

(2) Inspired from neuroscience, a Dual-Path Feed Forward Network (DPFFN) is introduced to emulate the parallel and cascaded information processing mechanisms in the biological visual cortex, and a Bionic Focal Vision (BFV) block is designed based on the LGF and DPFFN.

(3) Following the hierarchical design concept, a unified and efficient general pyramid backbone network family, called FViTs, is developed. Compared with other counterparts, FViTs demonstrate significant advantages in computational efficiency and scalability.

The remainder of this paper is organized as follows. Section 2 summarizes the related work of this paper in vision transformer and Gabor filter, respectively. The design process of FViTs is described in Section 3. Section 4 shows the experimental results of FViTs on various vision tasks. Section 5 is the conclusion.

\section{Related Work}

\subsection{Transformers for Vision}
The Transformer~\cite{vaswani2017attention} is originally conceived and implemented for machine translation and is regarded as a significant milestone in the field of Natural Language Processing (NLP)~\cite{wang2024speechx, wang2024viola}. Since 2020, a question has been raised: What happens when the Transformer is applied to the field of computer vision? Thereafter, substantial progress has been made. ViT~\cite{dosovitskiy2020image} is a groundbreaking endeavor that brings the transformer architecture into the domain of vision tasks. This innovative approach replaces convolutions entirely with transformers, showcasing superior performance compared with Convolutional Neural Networks (CNNs) across various vision benchmark tasks. Subsequently, a series of vision transformer variants~\cite{ren2023sg, chen2023cf, pan2023slide} have been proposed, offering new solutions for computer vision tasks. In the early stages of vision transformers development, researchers tended to attribute the excellent performance of vision transformers to the self-attention. Most of the research focuses on designing more efficient self-attention modules~\cite{liu2023efficientvit, chen2022mixformer, huang2023vision, li2022mvitv2, zhu2023biformer} and better integrating the advantages of self-attention and convolution~\cite{guo2022cmt, wang2022pvt, shi2023evit}. EViT~\cite{shi2023evit} is one of these works that designs a Bi-Fovea Self-Attention (BFSA) inspired by the visual characteristics of eagle vision. The objective of the BFSA is to combine the advantages of convolution and self-attention, thereby reducing the high computational complexity during self-attention operations. Meanwhile, other works~\cite{yu2022metaformer, rao2023gfnet, yan2024diffusion, garg2024gestformer} have provided different insights. These works demonstrate that excellent models can still be obtained by using spatial pooling layers or multi-layer perceptrons instead of self-attention, while keeping the structure of the vision transformer unchanged. This shows that self-attention is not indispensable in the transformer architecture. Besides self-attention, there are several other alternatives that can better accomplish computer vision tasks.

\subsection{2D Gabor Filter}

As a practical mathematical tool in computer graphics, the Gabor filter~\cite{gabor1946theory} has been widely used in the field of image processing. Its essence is to design a set of 2D Gabor functions to process the feature representations of images. This function is composed of the product of a sinusoidal plane wave function and a Gaussian kernel function, which gives the Gabor filter with direction selectivity and spatial frequency selectivity. The mathematical definition of the 2D Gabor function is defined as

\vspace{-0.3cm}

\begin{equation}\!\!\!\!
	\begin{aligned}
		g(x,y;\lambda ,\theta ,\psi ,\gamma ,\sigma ) = {e^{( - \frac{{{{x'}^2} + {\gamma ^2}{{y'}^2}}}{{2{\sigma ^2}}})}} \cdot {e^{(i(2\pi \frac{{x'}}{\lambda } + \psi ))}}
	\end{aligned}
	\label{eq:(1)}
\end{equation}

\vspace{-0.2cm}

\begin{equation}
	\begin{aligned}
		~~~~~~~~~~~~~~~~~~~~x' = x\cos \theta  + \sin \theta
	\end{aligned}
	\label{eq:(2)}
\end{equation}

\vspace{-0.4cm}

\begin{equation}
	\begin{aligned}
		~~~~~~~~~~~~~~~~~~y' =  - x\sin \theta  + y\cos \theta
	\end{aligned}
	\label{eq:(3)}
\end{equation}

\vspace{0.27cm}

\noindent{where}, $x$ and $y$ is the coordinates of the pixel positions, respectively. The $\lambda$ is the wavelength, which affects the sensitivity of the Gabor filter to specific frequencies. $\theta$ is the kernel orientation. $\psi$ is the phase offset, which is used to adjust the phase of the Gabor wavelet. $\gamma$ is the aspect ratio, which determines the shape of the Gabor wavelet. $\sigma$ is the bandwidth, which represents the variance of the Gaussian wavelet. The mainstream approaches~\cite{cai2024ultra, wang2024learnable, alkhalifah2024physics} employ the real part of the Gabor function to analyze image features, which can be expressed as

\begin{equation}\!\!\!\!\!\!
	\begin{aligned}
		g'(x,y;\lambda ,\theta ,\psi ,\gamma ,\sigma ) = {e^{( - \frac{{{{x'}^2} + {\gamma ^2}{{y'}^2}}}{{2{\sigma ^2}}})}} \cdot \cos (2\pi \frac{{x'}}{\lambda } + \psi )
	\end{aligned}
	\label{eq:(4)}
\end{equation}

\vspace{0.15cm}

Numerous researches have indicated that the dynamic response characteristics of the Gabor filter closely resemble the physiological traits of the biological visual system in vertebrates~\cite{fang2024gabor, shen2024intelligent}. In recent years, several researches~\cite{sun2024modeling, zhu2023learning} have tried to integrate Gabor filter as modulation process into deep convolutional neural networks. This integration aims to improve the ability to extract invariant information from images and enhance the interpretability of deep neural networks in the context of image analysis tasks.

\section{Approach}

\subsection{Overall Architecture}

\begin{figure*}[t]
	\centering         
	\includegraphics[width=\textwidth]{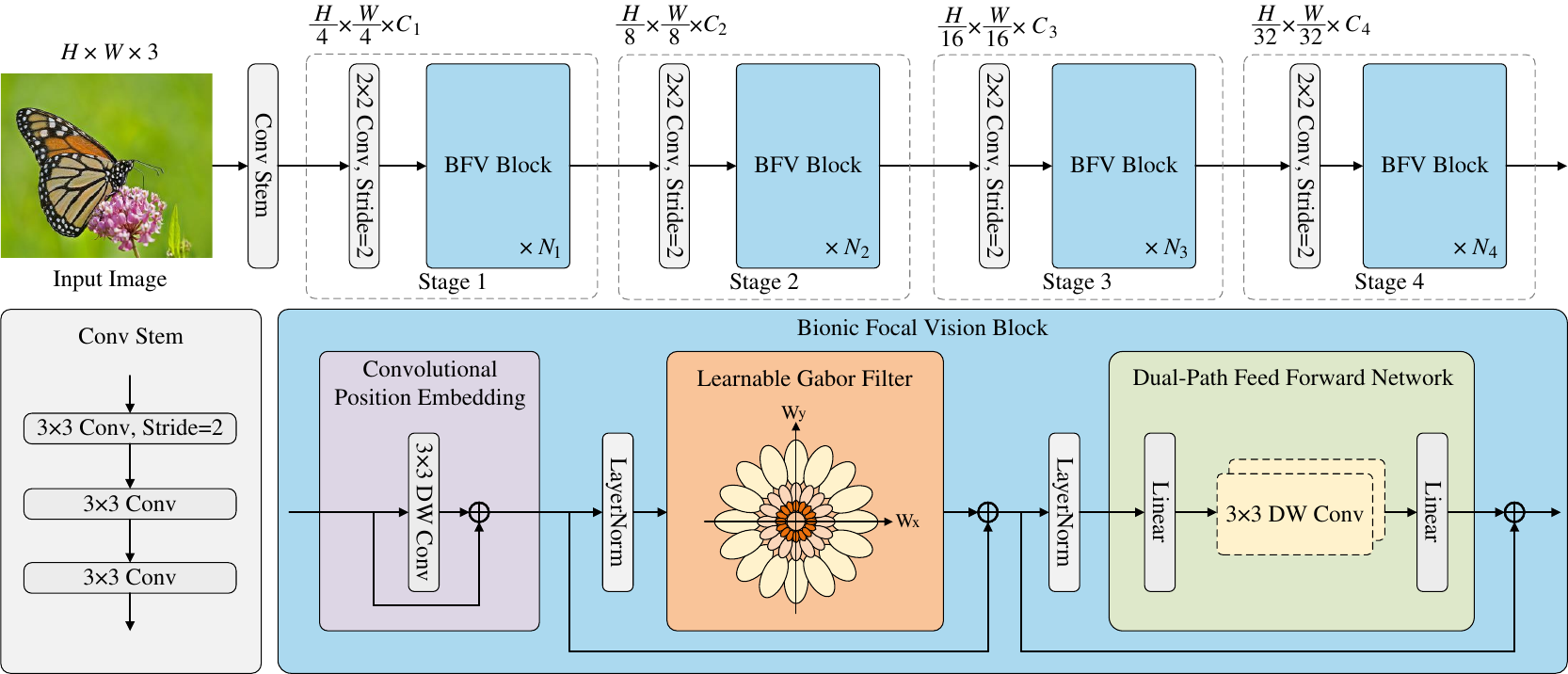}
	\caption{Overall architecture of the Focal Vision Transformer (FViT). FViT is composed of a convolutional stem and a pyramid structure with four stages. Each stage includes of a $2 \times 2$ convolution with stride 2 and multiple Bionic Focal Vision (BFV) blocks. The BFV block consists of a Convolutional Positional Embedding (CPE), a Learnable Gabor Filter (LGF) and a Dual-Path Feed Forward Network (DPFFN).}
	\label{fig:FViTs}
\end{figure*} 

Inspired by biological vision, a unified and efficient family of pyramid backbone networks called Focal Vision Transformers (FViTs) is developed. The overall pipeline of the proposed FViT is illustrated in Fig.~\ref{fig:FViTs}. It consists of a convolutional stem, several ${2 \times 2}$ convolutional layers and Bionic Focal Vision (BEV) blocks. To enable the FViT to be applied in high resolution dense prediction tasks, the mainstream hierarchical design concept is followed~\cite{li2023uniformer, shi2023evit, wang2022pvt}. Given an input image of size ${H \times W \times 3}$, it is first fed into the convolutional stem to extract low-level feature representations. This convolutional stem consists of three successive $3 \times 3$ convolutional layers, where the first convolutional layer is with a stride of 2 to stabilize
the training process of FViTs. Subsequently, these low-level feature representations are processed through the four stages of FViT to model the hierarchical feature representations of the target. From stage 1 to stage 4, the resolution of the output features is divided by factors 4, 8, 16, and 32, respectively, while the corresponding channel dimensions are increased to $C_1$, $C_2$, $C_3$, and $C_4$. Finally, in the image classification task, a normalization layer, an average pooling layer, and a fully connected layer are employed as classifiers to output the predictions.

\begin{figure}[t]
	\centering\small\resizebox{1.0\linewidth}{!}{
		\begin{tabular}{c}
			\hspace{-3mm}\includegraphics[width=1.0\textwidth]{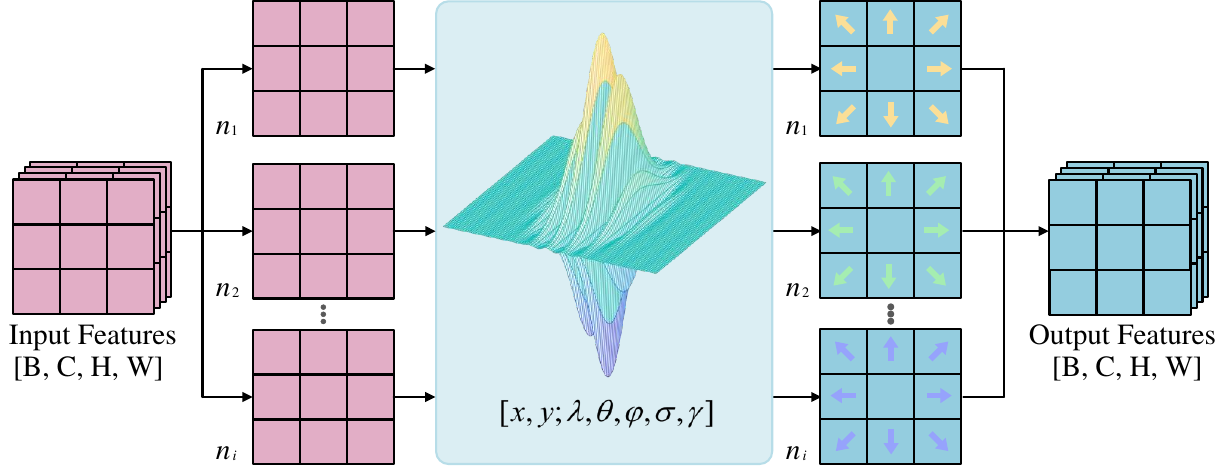}
	\end{tabular}}
	\caption{Illustration of the Learnable Gabor Filter (LGF)}
	\label{fig:FPA}
\end{figure}

\subsection{Bionic Focal Vision Block}

As the basic building units of the proposed FViTs, this BFV block consists of a Convolutional Position Embedding (CPE), a Learnable Gabor Filter (LGF) and a Dual-Path Feed Forward Network (DPFFN). It combines the advantages of convolutions and vision transformers for efficient modeling the feature representations of targets. The BFV block is defined as

\vspace{-0.2cm}

\begin{equation}
	\begin{aligned}
		~~~~~~~~~~~~~~~~~{\bf{X}} = {\rm{CPE}}({{\bf{X}}_{in}}) + {{\bf{X}}_{in}}
	\end{aligned}
	\label{eq:(5)}
\end{equation}

\vspace{-0.2cm}

\begin{equation}
	\begin{aligned}
		~~~~~~~~~~~~~~~~~{\bf{Y}} = {\rm{LGF}}({\rm{LN}}({\bf{X}})) + {\bf{X}}
	\end{aligned}
	\label{eq:(6)}
\end{equation}

\vspace{-0.27cm}

\begin{equation}
	\begin{aligned}
		~~~~~~~~~~~~~~~~~~{\bf{Z}} = {\rm{DPFFN}}({\rm{LN}}({\bf{Y}})) + {\bf{Y}}
	\end{aligned}
	\label{eq:(7)}
\end{equation}

\noindent{where} LN is the LayerNorm function, which is employed to normalize the feature tensors. Taking the first stage of FViTs as an example. An input feature tensor ${{\bf{X}}_{in}} \in {\mathbb{R}^{\frac{H}{4} \times \frac{W}{4} \times {C_1}}}$ is first processed using the CPE to introduce positional information to the feature tokens. Subsequently, these feature tokens are fed into the LGF to extract multi-scale and multi-orientation local features, thereby enabling the BFV to focus on the invariant feature representations of the targets. Finally, DPFFN is employed to enhance the ability of feature fusion and interaction for the BFV block. Compared with other advanced vision transformers~\cite{shi2023evit, wang2022pvt}, two key designs are used in the BFV block. (1) The LGF is employed to replace the self-attention in vision transformer blocks. (2) A novel DPFFN is introduced into the BFV block. These two designs are employed to reduce the computational complexity and memory cost of the FViT, while enhancing its scalability and generalization performance.

\begin{figure}[t]
	\centering\small\resizebox{1.0\linewidth}{!}{
		\begin{tabular}{c}
			\hspace{-3mm}\includegraphics[width=1.0\textwidth]{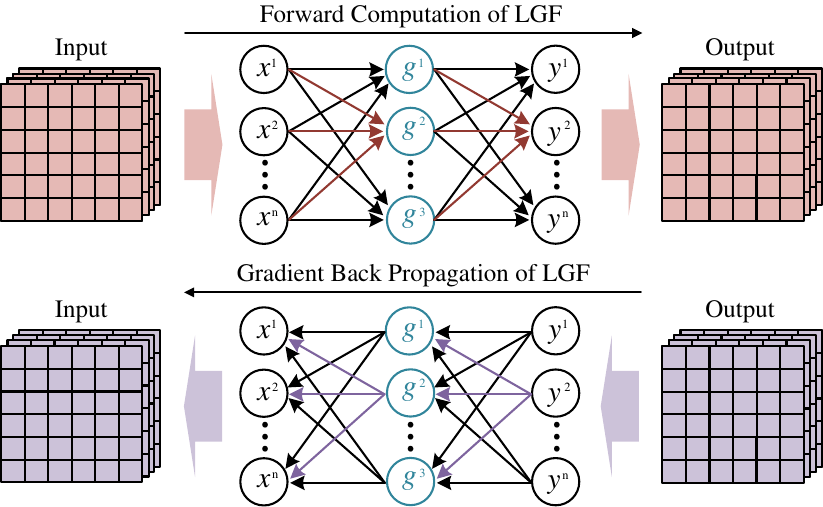}
	\end{tabular}}
	\caption{Illustration of the forward computation process and gradient backpropagation process of LGF.}
	\label{fig:FB}
\end{figure}

\subsection{Learnable Gabor Filter}

Fig.~\ref{fig:FPA} illustrates the computational process of LGF. The motivation behind it is to use Gabor filters to simulate the response of biological visual cells to input features, thereby capturing invariant feature representations to improve the interpretability and robustness of FViTs to complex features. However, as described in Equation~\ref{eq:(4)}, the Gabor function requires setting several hyperparameters, such as wavelength $\lambda$, kernel orientation $\theta$, and phase offset $\psi$. These hyperparameters are heavily relied on the prior knowledge and are sub-optimal. These parameter configurations directly affect the performance of the Gabor filter. For these reasons, a learnable Gabor filter is designed using convolution. Given an LGF with a kernel size of $k \times k$ defined as 

\begin{table*}[b]
	\caption{Four variants of FViTs for ImageNet-1K classification. $d_i$ represents the drop path rate, which is used to prevent overfitting of FViTs and to enhance its generalization ability. $r_i$ signifies the expansion ratio within the DPFFN of Stage \emph{i}. ${k_i}$ and ${p_i}$ are the kernel size and padding of the LGF, respectively. For ease of illustration, the size of input images of these four variants are all ${224^2}$.}
	
	\renewcommand{\arraystretch}{0.88}
	\setlength\tabcolsep{3.25pt}
	\centering
	\small
	\begin{tabular}{c|c|c|c|c|c}
		\toprule [0.125em]
		Output size & Layer Name & FViT-Tiny & FViT-Small & FViT-Base & FViT-Large \\ \midrule
		$112\times112$ & Conv Stem
		& $\begin{array}{c} 3\times3, 40, \text{stride}~2 \\ \left[3\times3, 40\right] \times 2 \end{array}$
		& $\begin{array}{c} 3\times3, 44, \text{stride}~2 \\ \left[3\times3, 44\right] \times 2 \end{array}$
		& $\begin{array}{c} 3\times3, 48, \text{stride}~2 \\ \left[3\times3, 48\right] \times 2 \end{array}$
		& $\begin{array}{c} 3\times3, 52, \text{stride}~2 \\ \left[3\times3, 52\right] \times 2 \end{array}$ \\ \midrule
		
		$56\times56$ & Convolution Layer & $2\times2$, $80$, stride $2$ & $2\times2$, $88$, stride $2$ & $2\times2$, $96$, stride $2$ & $2\times2$, $104$, stride $2$ \\ \midrule
		
		\begin{tabular}{c} Stage 1 \end{tabular} & \begin{tabular}{c}BFV block\end{tabular}
		& $\begin{bmatrix}\setlength{\arraycolsep}{1pt} \begin{array}{c}
				k_1$=$7, r_1$=$3.0 \\ p_1$=$3, d_1$=$0.05
		\end{array} \end{bmatrix} \times 3$
		& $\begin{bmatrix}\setlength{\arraycolsep}{1pt} \begin{array}{c}
				k_1$=$7, r_1$=$3.0 \\ p_1$=$3, d_1$=$0.05
		\end{array} \end{bmatrix} \times4$
		& $\begin{bmatrix}\setlength{\arraycolsep}{1pt} \begin{array}{c}
				k_1$=$7, r_1$=$3.0 \\ p_1$=$3, d_1$=$0.05
		\end{array} \end{bmatrix} \times 5$
		& $\begin{bmatrix}\setlength{\arraycolsep}{1pt} \begin{array}{c}
				k_1$=$7, r_1$=$3.0 \\ p_1$=$3, d_1$=$0.05
		\end{array} \end{bmatrix} \times 5$ \\ \midrule
		
		$28\times28$ & Convolution Layer & $2\times2$, $160$, stride $2$ & $2\times2$, $176$, stride $2$ & $2\times2$, $192$, stride $2$ & $2\times2$, $208$, stride $2$ \\ \midrule 
		\begin{tabular}{c} Stage 2 \end{tabular} & \begin{tabular}{c}BFV block \end{tabular} 
		& $\begin{bmatrix}\setlength{\arraycolsep}{1pt} \begin{array}{c}
				k_1$=$5, r_1$=$3.5 \\ p_1$=$2, d_1$=$0.05
		\end{array} \end{bmatrix} \times 3$
		& $\begin{bmatrix}\setlength{\arraycolsep}{1pt} \begin{array}{c}
				k_1$=$5, r_1$=$3.5 \\ p_1$=$2, d_1$=$0.05
		\end{array} \end{bmatrix} \times4$
		& $\begin{bmatrix}\setlength{\arraycolsep}{1pt} \begin{array}{c}
				k_1$=$5, r_1$=$3.5 \\ p_1$=$2, d_1$=$0.05
		\end{array} \end{bmatrix} \times 5$
		& $\begin{bmatrix}\setlength{\arraycolsep}{1pt} \begin{array}{c}
				k_1$=$5, r_1$=$3.5 \\ p_1$=$2, d_1$=$0.05
		\end{array} \end{bmatrix} \times 5$ \\ \midrule
		
		$14\times14$ & Convolution Layer & $2\times2$, $320$, stride $2$ & $2\times2$, $352$, stride $2$ & $2\times2$, $384$, stride $2$ & $2\times2$, $416$, stride $2$ \\ \midrule
		
		\begin{tabular}{c} Stage 3 \end{tabular} & \begin{tabular}{c}BFV block
		\end{tabular} 
		& $\begin{bmatrix}\setlength{\arraycolsep}{1pt} \begin{array}{c}
				k_1$=$3, r_1$=$4.0 \\ p_1$=$1, d_1$=$0.05
		\end{array} \end{bmatrix} \times 12$
		& $\begin{bmatrix}\setlength{\arraycolsep}{1pt} \begin{array}{c}
				k_1$=$3, r_1$=$4.0 \\ p_1$=$1, d_1$=$0.05
		\end{array} \end{bmatrix} \times 20$
		& $\begin{bmatrix}\setlength{\arraycolsep}{1pt} \begin{array}{c}
				k_1$=$3, r_1$=$4.0 \\ p_1$=$1, d_1$=$0.05
		\end{array} \end{bmatrix} \times 28$
		& $\begin{bmatrix}\setlength{\arraycolsep}{1pt} \begin{array}{c}
				k_1$=$3, r_1$=$4.0 \\ p_1$=$1, d_1$=$0.05
		\end{array} \end{bmatrix} \times 36$ \\ \midrule
		
		$7\times7$ & Convolution Layer & $2\times2$, $640$, stride $2$ & $2\times2$, $704$, stride $2$ & $2\times2$, $768$, stride $2$ & $2\times2$, $832$, stride $2$ \\ \midrule
		
		\begin{tabular}{c} Stage 4 \end{tabular} & \begin{tabular}{c}BFV block
		\end{tabular} 
		& $\begin{bmatrix}\setlength{\arraycolsep}{1pt} \begin{array}{c}
				k_1$=$3, r_1$=$4.0 \\ p_1$=$1, d_1$=$0.05
		\end{array} \end{bmatrix} \times 3$
		& $\begin{bmatrix}\setlength{\arraycolsep}{1pt} \begin{array}{c}
				k_1$=$3, r_1$=$4.0 \\ p_1$=$1, d_1$=$0.05
		\end{array} \end{bmatrix} \times 4$
		& $\begin{bmatrix}\setlength{\arraycolsep}{1pt} \begin{array}{c}
				k_1$=$3, r_1$=$4.0 \\ p_1$=$1, d_1$=$0.05
		\end{array} \end{bmatrix} \times 5$
		& $\begin{bmatrix}\setlength{\arraycolsep}{1pt} \begin{array}{c}
				k_1$=$3, r_1$=$4.0 \\ p_1$=$1, d_1$=$0.05
		\end{array} \end{bmatrix} \times 5$ \\ \midrule
		
		$1\times1$ & Projection & \multicolumn{4}{c}{$1\times1$, $1280$} \\ \midrule
		$1\times1$ & Classifier & \multicolumn{4}{c}{Fully Connected Layer, $1000$} \\ \midrule
		\multicolumn{2}{c|}{Params} & $11.28$ M & $23.59$ M & $43.15$ M & $60.15$ M \\ \midrule
		\multicolumn{2}{c|}{FLOPs} & $1.94$ G & $3.98$ G & $7.21$ G & $10.18$ G \\
		\bottomrule[0.125em]
	\end{tabular}
	\label{tab:arch_variants_of_FViT}
\end{table*}

\begin{algorithm}[t]
	Input: ${{\bf{F}}} \in {\mathbb{R}^{H \times W \times C}}$
	
	Parameters: index $x$ and $y$; wavelength $\lambda$, kernel orientation $\theta$, bandwidth $\sigma$, phase offset $\psi$, aspect ratio $\gamma$, kernel size $k$, kernel of LGF $kernel = [~~]$.
	
	\# Define the Learnable Gabor Filter (LGF) function 
	
	def LGF:
	
	~~~~~~$g'(x,y;\lambda ,\theta ,\psi ,\gamma ,\sigma ) = {e^{( - \frac{{{{x'}^2} + {\gamma ^2}{{y'}^2}}}{{2{\sigma ^2}}})}} \cdot \cos (2\pi \frac{{x'}}{\lambda } + \psi )$
	
	~~~~~~for~~$i$~~in~~$k$:
	
	~~~~~~~~~~~~for~~$j$~~in~~$k$:
	
	~~~~~~~~~~~~~~~~~~$kernel[i][j] = g'(x,y;\lambda ,\theta ,\psi ,\gamma ,\sigma )$
	
	~~~~~~~~~~~~end
	
	~~~~~~end
	
	return Conv2d($weight = kernel$)
	
	\caption{Learnable Gabor Filter algorithm.}
	\label{alg:FPA}
\end{algorithm}

\begin{equation}\!\!\!\!\!\!\!\!\!\!
	\begin{aligned}
		~~~~~~~~LGF = \left[ {\begin{array}{*{20}{c}}
				{g_w^{(0,0)}}&{g_w^{(1,0)}}& \cdots &{g_w^{(k,0)}}\\
				{g_w^{(0,1)}}&{g_w^{(1,1)}}& \cdots &{g_w^{(k,1)}}\\
				\vdots & \vdots & \ddots & \vdots \\
				{g_w^{(0,k)}}&{g_w^{(1,k)}}& \cdots &{g_w^{(k,k)}}
		\end{array}} \right]
	\end{aligned}
	\label{eq:(8)}
\end{equation}

\noindent{where} the $g_\omega ^{(i,j)} = g'(x,y;\lambda ,\theta ,\psi ,\gamma ,\sigma )$, and $\lambda$, $\theta$, $\psi$, $\gamma$, $\sigma$ are all trainable parameters. Taking the wavelength $\lambda$ as an example, the gradient of $g_\omega ^{(i,j)}$ with respect to $\lambda$ can be expressed as

\begin{equation}\!\!\!
	\begin{aligned}
		~~~~~~~~\frac{{\partial g'}}{{\partial \lambda }} = 2\pi \frac{{x'}}{{{\lambda ^2}}}{e^{( - \frac{{{{x'}^2} + {\lambda ^2}{{y'}^2}}}{{2{\sigma ^2}}})}}\sin (2\pi \frac{{x'}}{\lambda } + \psi )
	\end{aligned}
	\label{eq:(9)}
\end{equation}

\vspace{0.27cm}

Subsequently, the parameters in the LGF are iteratively learned and updated using the optimizer and gradient backpropagation algorithm during the training phase of FViT. The implementation details of the forward computation process and the gradient backpropagation process of the LGF are shown in Fig.~\ref{fig:FB}. The proposed LGF is used as an alternative to self-attention. Compared with self-attention, the computational complexity and memory cost of LGF are linear with respect to the number of feature tokens, thereby enhancing the competitive advantage of vision transformers. This means that FViT can be built deeper and wider, which is beneficial for extracting richer semantic information of the targets. Algorithm~\ref{alg:FPA} summarizes the calculation process of the LGF.

\subsection{Dual-Path Feed Forward Network}

\begin{figure}[t]
	\centering\small\resizebox{1.0\linewidth}{!}{
		\begin{tabular}{c}
			\hspace{-3mm}\includegraphics[width=1.0\textwidth]{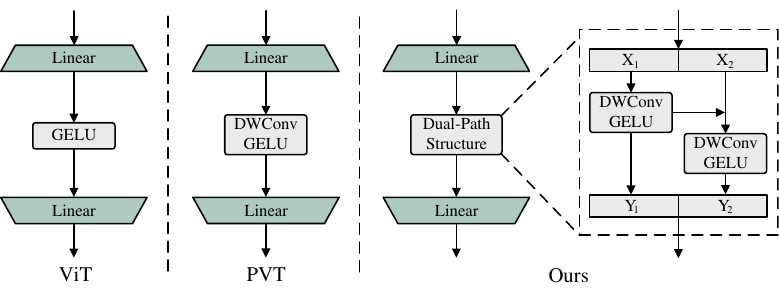}
	\end{tabular}}
	\caption{Visualization of the Dual-Path Feed Forward Network (DPFFN).}
	\label{fig:DPFFN}
\end{figure}

As a key component in vision transformers, the feed forward network is employed to integrate the global dependencies between different feature representations through nonlinear mapping. However, the feed forward network lacks sensitivity to local features. The common approach is to introduce convolution operations between the two fully connected layers, or use $1 \times 1$ convolutions to replace the fully connected layers. Nevertheless, these approaches are regarded as inefficient. To this end, inspired by the information processing mechanisms of the biological visual cortex, it is believed that an efficient feed forward network should satisfy the two conditions, including the hierarchical structure and parallel processing. Therefore, a novel Dual-Path Feed Forward Network (DPFFN) is designed, which is illustrated in Fig.~\ref{fig:DPFFN}. In DPFFN, the input features are partitioned into two groups, and the output features from the preceding group are subsequently transmitted to the next group of filters. This process involves the information flow path with another set of input features. The DPFFN has the characteristics of both hierarchical structural and parallel processing, which can help expand the receptive field within each network layer, thereby enhancing the multi-scale feature representation of networks at more complex levels.

\subsection{Architecture Variants of FViTs}

FViTs follow the mainstream hierarchical design concept~\cite{li2023uniformer, shi2023evit, wang2022pvt} and comprise four variations: FViT-Tiny, FViT-Small, FViT-Base, and FViT-Large. These variants consist of four stages, each with different number of BFV blocks and hidden feature dimensions. Specifically, $2 \times 2$ convolutions with a stride of 2 are employed to connect different stages for patch embedding. This process doubles the dimensions of feature maps and halves the spatial size before they are fed into the next stage. Therefore, each stage can output features of different sizes and dimensions, thereby obtaining rich hierarchical feature representations of targets. The configuration details of the FViTs are shown in Table~\ref{tab:arch_variants_of_FViT}. To facilitate comparison with other advanced visual neural networks, the input image resolution of FViT-Tiny, FViT-Small, FViT-Base and FViT-Large is set to ${224^2}$.

\section{Experiments}

In this section, experiments are conducted on FViTs in a series of mainstream computer vision tasks, including image classification (Sec.~{4.1}), object detection (Sec.~{4.2}), and semantic segmentation (Sec.~{4.3}). Specifically, FViTs are first trained from scratch on the ImageNet-1K~\cite{russakovsky2015imagenet} dataset for image classification to obtain pre-training parameters. Subsequently, the pre-training parameters of FViTs are fine-tuned through transfer learning on object detection and semantic segmentation tasks, which are used to evaluate the generalization performance of FViTs. In addition, ablation experiments on FViTs are conducted in Section 4.4 to validate the effectiveness of LGF and DPFFN.

\begin{table}[t]
	\caption{ImageNet-1K classification results of FViTs. Similar CNNs, Transformers and MLPs are grouped together based on their parameters and classification performance.}
	\begin{center}
		\resizebox{1.0\linewidth}{!}{
			\begin{tabular}{l|c|c|c|c}
				\toprule[0.125em]
				\multirow{1}{*}{Model} & Resolution & FLOPs (G)& Params (M)& Top-1 Acc (\%)\\
				
				\midrule
				ResNet-18~\cite{he2016deep} & $224^2$ & 1.8 & 11.7 & 69.8 \\
				PVT-T~\cite{wang2021pyramid} & $224^2$ & 1.9 & 13.2 & 75.1 \\
				EfficientVMamba-T~\cite{pei2024efficientvmamba} & $224^2$ & {0.8} & {6.0} & 76.5 \\
				ResMLP-S12~\cite{touvron2022resmlp} & $224^2$ & 3.0 & 15.4 & 76.6 \\
				ResT-Lite~\cite{zhang2021rest} & $224^2$ & 1.4 & 10.5 & 77.2 \\
				PoolFormer-S12~\cite{yu2022metaformer} & $224^2$ & 1.9 & 12.0 & 77.2 \\
				RegNetY-1.6~\cite{radosavovic2020designing} & $224^2$ & 1.6 & 11.2 & 78.0 \\
				PVTv2-B1~\cite{wang2022pvt} & $224^2$ & 2.1 & 13.1 & 78.7 \\
				EfficientVMamba-S~\cite{pei2024efficientvmamba} & $224^2$ & 1.3 & 11.0 & 78.7 \\
				\rowcolor{mygray} FViT-Tiny & $224^2$ & 1.9 & 11.3 & 79.1 \\
				
				\midrule
				ResNet-50~\cite{he2016deep} & $224^2$ & 4.1 & 25.6 & 76.2 \\
				RegNetY-4.0~\cite{radosavovic2020designing} & $224^2$ & 4.0 & {20.6} & 79.4 \\
				ResMLP-S24~\cite{touvron2022resmlp} & $224^2$ & 6.0 & 30.0 & 79.4 \\
				PVT-S~\cite{wang2021pyramid} & $224^2$ & 3.8 & 24.5 & 79.8 \\
				PoolFormer-S24~\cite{yu2022metaformer} & $224^2$ & 3.5 & 21.0 & 80.3 \\
				Swin-T~\cite{liu2021swin} & $224^2$ & 4.5  & 28.3 & 81.3 \\
				T2T-14~\cite{yuan2021tokens} & $224^2$ & 5.2 & 21.5 & 81.5 \\
				LIT-S~\cite{pan2022less} & $224^2$ & 4.1 & 27.0 & 81.5 \\
				CrossFormer-T~\cite{wang2023crossformer++} & $224^2$ & 2.9 & 27.8 & 81.5 \\
				ResT-Base~\cite{zhang2021rest} & $224^2$ & 4.3 & 30.3 & 81.6 \\
				PVTv2-B2~\cite{wang2022pvt} & $224^2$ & 4.0 & 25.4 & 82.0 \\
				ConvNeXt-T~\cite{liu2022convnet} & $224^2$ & 4.5 & 28.0 & 82.1 \\
				\rowcolor{mygray} FViT-Small & $224^2$ & {4.0} & 23.6 & {82.3}\\
				
				\midrule
				ResNet-101~\cite{he2016deep} & $224^2$ & 7.9 & 44.7 & 77.4 \\
				RegNetY-8.0~\cite{radosavovic2020designing} & $224^2$ & 8.0 & 39.2 & 79.9 \\
				PVT-M~\cite{wang2021pyramid} & $224^2$ & 6.7 & 44.2 & 81.2 \\
				PoolFormer-S36~\cite{yu2022metaformer} & $224^2$ & 5.1 & 31.0 & 81.4 \\
				EfficientVMamba-B~\cite{pei2024efficientvmamba} & $224^2$ & 4.0 & {33.0} & 81.8 \\
				T2T-19~\cite{yuan2021tokens} & $224^2$ & 9.8 & 39.2 & 81.9 \\
				CrossFormer-S~\cite{wang2023crossformer++} & $224^2$ & 4.9 & 30.7 & 82.5 \\
				MSVMamba-M~\cite{shi2024multi} & $224^2$ & {4.6} & {33.0} &  82.8 \\
				LIT-M~\cite{pan2022less} & $224^2$ & 8.6 & 48.0 & 83.0 \\
				Swin-S~\cite{liu2021swin} & $224^2$ & 8.7 & 49.6 & 83.0 \\
				ConvNeXt-S~\cite{liu2022convnet} & $224^2$ & 8.7 & 50.0 & 83.1 \\
				PVTv2-B3~\cite{wang2022pvt} & $224^2$ & 6.9 & 45.2 & 83.2 \\
				\rowcolor{mygray} FViT-Base & $224^2$ & 7.2 & 43.2 & {83.4} \\
				
				\midrule
				ResNet-152~\cite{he2016deep} & $224^2$ & 11.6 & 60.2 & 78.3 \\
				RegNetY-16~\cite{radosavovic2020designing} & $224^2$ & 15.9 & 83.6 & 80.4 \\
				PVT-L~\cite{wang2021pyramid} & $224^2$ & 9.8 & 61.4 & 81.7 \\
				PoolFormer-M36~\cite{yu2022metaformer} & $224^2$ & 9.0 & 56.0 & 82.1 \\
				T2T-24~\cite{yuan2021tokens} & $224^2$ & 15.0 & 64.1 & 82.2 \\
				LIT-B~\cite{pan2022less} & $224^2$ & 15.0 & 86.0 & 83.4 \\
				CrossFormer-B~\cite{wang2023crossformer++} & $224^2$ & 9.5 & 52.0 & 83.4 \\
				Swin-B~\cite{liu2021swin} & $224^2$ & 15.4 & 87.8 & 83.5 \\
				PVTv2-B4~\cite{wang2022pvt} & $224^2$ & 10.1 & 62.6 & 83.6 \\
				\rowcolor{mygray} FViT-Large & $224^2$ & {10.2} & {60.2} & {83.8}\\
				\bottomrule[0.125em]
		\end{tabular}}
	\end{center}
	\label{tab:class_results}
\end{table}

\begin{table*}[t]
	\caption{A comparative performance analysis is carried out for object detection (in the left group) and instance segmentation (in the right group) utilizing the COCO 2017 validation dataset. Each model serves as a vision backbone and is subsequently integrated into the RetinaNet~\cite{lin2017focal} and Mask R-CNN~\cite{he2017mask} frameworks.}
	\begin{center}
		\small\resizebox{1.0\linewidth}{!}{
			\begin{tabular}{l|ccccccc|ccccccc}
				\toprule[0.125em]
				\multirow{2}{*}{Backbone} & \multicolumn{7}{c|}{RetinaNet} & \multicolumn{7}{c}{Mask R-CNN}  \\
				& Params (M) &  $mAP$ & $AP_{50}$  & $AP_{75}$ & $AP_S$ &  $AP_M$ & $AP_L$ & Params (M) & $mAP^b$ & $AP^b_{50}$  & $AP^b_{75}$ & $mAP^m$ &  $AP^m_{50}$ & $AP^m_{75}$ \\
				\midrule
				ResNet-50~\cite{he2016deep} & 37.7 & 36.3 & 55.3 & 38.6 & 19.3 & 40.0 & 48.8 & 44.2 & 38.0 & 58.6 & 41.4 & 34.4 & 55.1 & 36.7 \\
				PVT-S~\cite{wang2021pyramid} & 34.2 & 40.4 & 61.3 & 43.0 & 25.0 & 42.9 & 55.7 & 44.1 & 40.4 & 62.9 & 43.8 & 37.8 & 60.1 & 40.3 \\
				Swin-T~\cite{liu2021swin} & 38.5 & 42.0 & 63.0 & 44.7 & 26.6 & 45.8 & 55.7 & 47.8 & 42.2 & 64.6 & 46.2 & 39.1 & 61.6 & 42.0 \\
				ResT-Base~\cite{zhang2021rest} & 40.5 & 42.0 & 63.2 & 44.8 & 29.1 & 45.3 & 53.3 & 49.8 & 41.6 & 64.9 & 45.1 & 38.7 & 61.6 & 41.4 \\
				DAT-T~\cite{xia2022vision} & 38.0 & 42.8 & 64.4 & 45.2 & 28.0 & 45.8 & 57.8 & 48.0 & 44.4 & 67.6 & 48.5 & 40.4 & 64.2 & 43.1 \\
				CMT-S~\cite{guo2022cmt} & 44.0 & 44.3 & 65.5 & 47.5 & 27.1 & 48.3 & 59.1 & 45.0 & 44.6 & 66.8 & 48.9 & 40.7 & 63.9 & 43.4 \\
				PVTv2-B2~\cite{wang2022pvt} & 35.1 & 44.6 & 65.6 & 47.6 & 27.4 & 48.8 & 58.6 & 45.0 & 45.3 & 67.1 & 49.6 & 41.2 & 64.2 & 44.4 \\
				CrossFormer-S~\cite{wang2023crossformer++} & 40.8 & 44.4 & 65.8 & 47.4 & 28.2 & 48.4 & 59.4 & 50.2 & 45.4 & 68.0 & 49.7 & 41.4 & 64.8 & 44.6 \\
				\rowcolor{mygray}
				FViT-Small & {33.7} & {44.7} & {65.8} & {47.9} & {28.3} & {48.8} & {59.5} & {43.9} & {45.6} & {68.1} & {50.0} & {41.5} & {64.9} & {44.7} \\
				\midrule
				\midrule
				ResNet-101~\cite{he2016deep} & 56.7 & 38.5 & 57.8 & 41.2 & 21.4 & 42.6 & 51.1 & 63.2 & 40.4 & 61.1 & 44.2 & 36.4 & 57.7 & 38.8 \\
				PVT-M~\cite{wang2021pyramid} & 53.9 & 41.9 & 63.1 & 44.3 & 25.0 & 44.9 & 57.6 & 63.9 & 42.0 & 64.4 & 45.6 & 39.0 & 61.6 & 42.1 \\
				Swin-S~\cite{liu2021swin} & 59.8 & 44.5 & 65.7 & 47.5 & 27.4 & 48.0 & 59.9 & 69.1 & 44.8 & 66.6 & 48.9 & 40.9 & 63.4 & 44.2 \\
				DAT-S~\cite{xia2022vision} & 60.0 & 45.7 & 67.7 & 48.5 & 30.5 & 49.3 & 61.3 & 69.0 & 47.1 & 69.9 & 51.5 & 42.5 & 66.7 & 45.4 \\
				ScalableViT-B~\cite{yang2022scalablevit} & 85.0 & 45.8 & 67.3 & 49.2 & 29.9 & 49.5 & 61.0 & 95.0 & 46.8 & 68.7 & 51.5 & 42.5 & 65.8 & 45.9 \\
				PVTv2-B3~\cite{wang2022pvt} & 55.0 & 45.9 & 66.8 & 49.3 & 28.6 & 49.8 & 61.4 & 64.9 & 47.0 & 68.1 & 51.7 & 42.5 & 65.7 & 45.7 \\
				CrossFormer-B~\cite{wang2023crossformer++} & 62.1 & 46.2 & 67.8 & 49.5 & 30.1 & 49.9 & 61.5 & 71.5 & 47.2 & 69.9 & 52.0 & 42.7 & 66.6 & 46.2 \\
				\rowcolor{mygray}
				FViT-Base & {51.4} & {46.4} & {68.3} & {49.6} & {30.2} & {50.7} & {61.8} & {61.8} & {47.3} & {69.6} & {52.2} & {42.9} & {66.5} & {46.3} \\    
				\bottomrule[0.125em]
		\end{tabular}}
	\end{center}
	\label{tab:coco2017}
\end{table*}

\subsection{Image Classification on ImageNet-1k}

\noindent\textbf{Settings.} In this section, ImageNet-1K~\cite{russakovsky2015imagenet} is used as the dataset for image classification. This dataset contains 1000 classes with approximately 1.33 M images. Among these, 1.28 M images are used for training, and the remaining 50,000 images are served as validation data. For fairness, the same training strategy as used in PVTv2~\cite{wang2022pvt} and CrossFormer~\cite{wang2023crossformer++} is followed. Specifically, AdamW is selected as the parameter optimizer and the weight decay is set to 0.05. All models are trained for 300 epochs, and the initial learning rate is set to 0.001, followed by cosine decay. The same data augmentation techniques are employed as CrossFormer~\cite{wang2023crossformer++}, including Random Flip, Random Crop, Random Erase, CutMix, Mixup, and Label Smoothing.

\begin{figure}[t]
	\centering\small\resizebox{1.0\linewidth}{!}{
		\begin{tabular}{c}
			\hspace{-3mm}\includegraphics[width=1.0\textwidth]{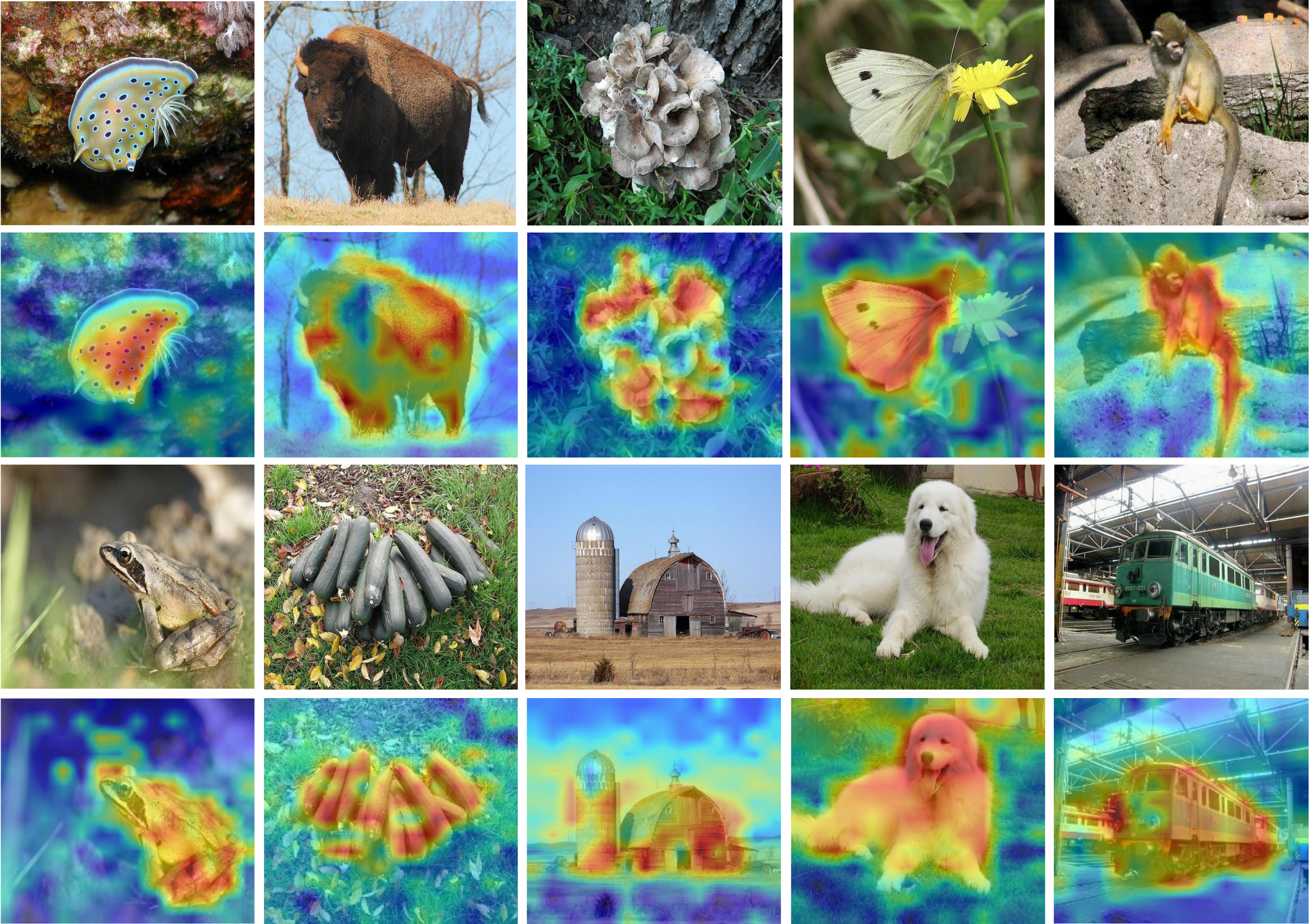}
	\end{tabular}}
	\caption{Visualization of the features modelled by FViTs. Regions with higher values indicate that they received more attention during the forward computation process of FViTs.}
	\label{fig:visual}
\end{figure}

\noindent\textbf{Results.} Table~\ref{tab:class_results} shows the performance of FViTs on the ImageNet-1K classification task. To facilitate comparison, similar networks are grouped based on their model parameters and performance. The numerical results indicate that FViTs demonstrate significant competitive advantages with similar model parameters and computational costs. Specifically, FViT-Tiny and FViT-Small achieve 79.1\% and 82.3\% classification accuracies at small model scales. Compared with other advanced networks, the classification accuracy of FViT-Tiny and FViT-Small is higher than other networks by 0.4\% and 0.2\%, respectively. At larger parameter scales, FViT-Base and FViT-Large maintain significant performance advantages over the other networks. The performance of FViT-Base and FViT-Large achieved 83.4\% and 83.8\% accuracy in the ImageNet-1K classification task, respectively. Under the same settings, FViT-Large achieves 0.3\% and 0.2\% performance gains compared with Swin-B~\cite{liu2021swin} and PVTv2-B4~\cite{wang2022pvt}, respectively. In addition, two advanced baselines without self-attention, ResMLP~\cite{touvron2022resmlp} and PoolFormer~\cite{yu2022metaformer}, are selected for comparison. The four variants of FViTs all demonstrate superior performance compared with ResMLP~\cite{touvron2022resmlp} and PoolFormer~\cite{yu2022metaformer}. For an in-depth analysis of the FViTs, the features modeled by the FViTs are visualized using heatmaps. As shown in Fig.~\ref{fig:visual}, regions with higher values indicate that they have received more attention during the forward computation of the network, reflecting the importance of the features. From the distribution of values in these heat maps, it can be seen that FViTs are able to effectively focus on foreground targets of interest while suppressing unnecessary background information. This characteristic enables FViTs to demonstrate excellent performance in handling visually diverse and complex tasks.

\subsection{Object Detection and Instance Segmentation}

\noindent\textbf{Settings.} Object detection and instance segmentation experiments are conducted on FViTs using the COCO 2017~\cite{lin2014microsoft} dataset. The COCO 2017 dataset comprises 80 classes, 118K training images, 5K validation images and 20K test images. In the experiments, RetinaNet~\cite{lin2017focal} and Mask R-CNN~\cite{he2017mask} are used as benchmarks to evaluate the robustness and generalization performance of FViTs. Specifically, FViT-Small and FViT-Base are employed as the backbone and then plugged into the RetinaNet and Mask R-CNN frameworks. Prior to training, FViT-Small and FViT-Base are initialized with pre-trained parameters from ImageNet-1K, while the remaining layers are randomly initialized. To ensure fairness, the same experimental configuration as used in PVTv2~\cite{wang2022pvt} is followed. Specifically, the shorter side of the input image is set to 800, with the longer side is allowed to be a maximum of 1333. The AdamW is selected as the parameter optimizer, and the training schedule is set to $1 \times 12$ epochs. The weight decay rate is set to 0.05, and the initial learning rate is set to 0.0001.

\begin{figure}[t]
	\centering\small\resizebox{1.0\linewidth}{!}{
		\begin{tabular}{c}
			\hspace{-3mm}\includegraphics[width=1.0\textwidth]{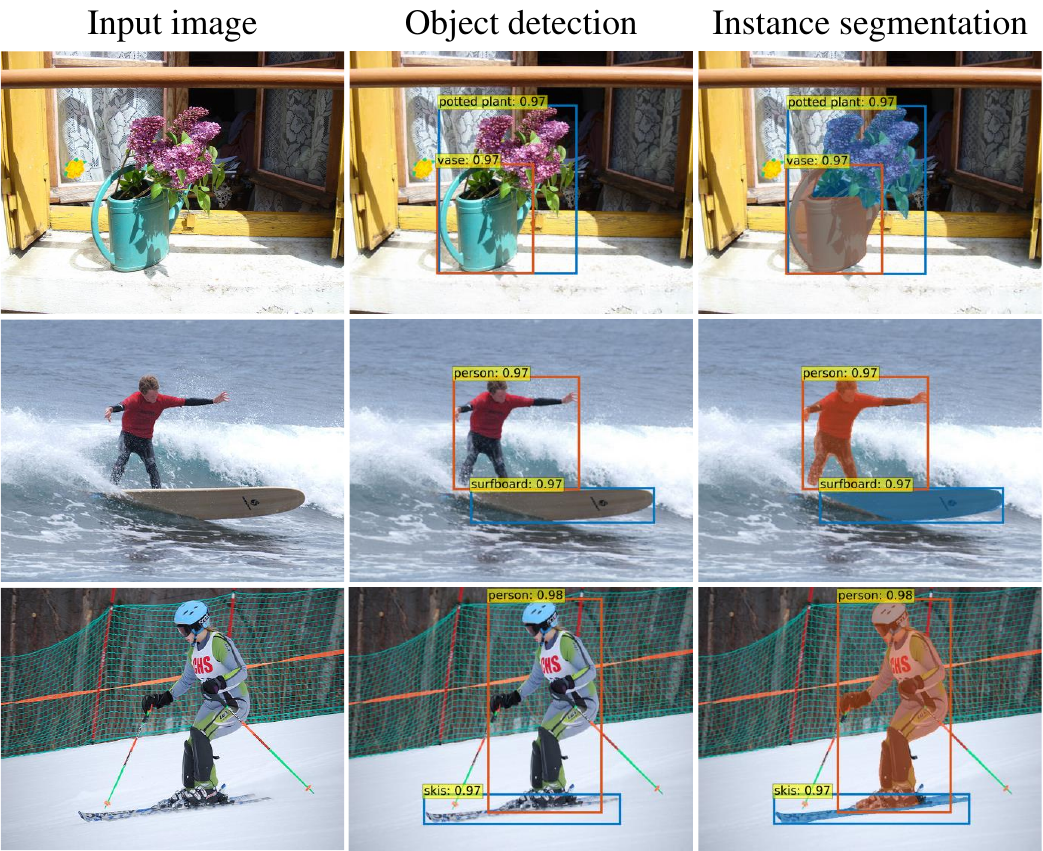}
	\end{tabular}}
	\caption{Qualitative results of FViTs for object detection and instance segmentation on the COCO 2017 validation set.}
	\label{fig:Object_instance}
\end{figure}

\noindent\textbf{Results.} Table~\ref{tab:coco2017} shows the performance comparison of FViTs with DAT~\cite{xia2022vision}, PVTv2~\cite{wang2022pvt} and CrossFormer~\cite{wang2023crossformer++} for object detection and instance segmentation on COCO 2017 validation dataset. In the RetinaNet framework, the mean Average Precision ($mAP$), Average Precision at 50\% and 75\% IoU thresholds ($A{P_{50}}$, $A{P_{75}}$), and Small, Medium, Large Average Precision ($A{P_{S}}$, $A{P_{M}}$, and $A{P_{L}}$) are used as the evaluation metrics to evaluate model performance. The numerical results show that FViTs exhibit superior performance compared with other advanced networks. Specifically, the average accuracies of FViT-Small and FViT-Base are at least 1.7\% higher than DAT-T~\cite{xia2022vision} and DAT-S~\cite{xia2022vision}, and they outperform the advanced CrossFormer-S~\cite{wang2023crossformer++} and CrossFormer-B~\cite{wang2023crossformer++} by 0.3\% and 0.2\%, respectively. In the Mask R-CNN framework, the mean bounding-box Average Precision ($mA{P^b}$), mean mask Average Precision ($mA{P^m}$), and $A{P^b}$ and $A{P^m}$ at different IoU thresholds (50\%, 75\%) are used to evaluate the performance of the networks. In terms of $mAP^m$ metric, FViT-Small and FViT-Base outperform PVTv2-B2~\cite{wang2022pvt}~\cite{xia2022vision} and PVTv2-B3~\cite{wang2022pvt} by 0.3\% and 0.4\%, respectively. These results indicate that the performance of FViT-Small and FViT-Base significantly outperforms the other networks, achieving a better trade-off between model parameters and accuracy. Furthermore, Fig.~\ref{fig:Object_instance} shows the predicted results of FViTs for object detection and instance segmentation on the COCO 2017 validation set. The robust qualitative results demonstrate the effectiveness of the FViTs.

\subsection{Semantic Segmentation on ADE20K}

\noindent\textbf{Settings.} The semantic segmentation experiments are conducted for FViTs on ADE20K~\cite{zhou2017scene} dataset. This ADE20K dataset comprises 150 different semantic classes, with about 20K training images, 2K validation images and 3K test images. To facilitate comparison with other networks, Semantic FPN~\cite{kirillov2019panoptic} is used as the benchmark. FViT-Small and FViT-Base are used as backbones and plugged into the semantic FPN to evaluate their performance in semantic segmentation tasks. For fairness, the same training strategy as PVTv2~\cite{wang2022pvt} and CrossFormer~\cite{wang2023crossformer++} is adopted for comparison with other advanced networks. Specifically, the AdamW is selected as the parameter optimizer and the learning rate is set to 1e-4. The learning rate is decayed by following the polynomial schedule with an exponent of 0.9, and the total number of training iterations is set to 80k.

\begin{table}[t]
	\caption{The performance comparison of FViTs with other backbone networks for semantic segmentation on the ADE20K dataset.
	}
	\begin{center}
		\resizebox{1.0\linewidth}{!}{
			\begin{tabular}{l|>{\centering\arraybackslash}m{1.5cm}|>{\centering\arraybackslash}m{1.5cm}|>{\centering\arraybackslash}m{1.5cm}}
				\toprule[0.125em]
				Backbone & Params (M) & FLOPs (G) & mIoU (\%) \\
				\midrule
				ResNet-50~\cite{he2016deep} & 28.5 & 45.6 & 36.7 \\
				PVT-S~\cite{wang2021pyramid} & 28.2 & {44.5} & 39.8 \\
				Swin-T~\cite{liu2021swin} & 32.0 & {46.0} & 41.5 \\
				LITv2-S~\cite{pan2022fast} & 31.0 & {41.0} & 44.3 \\
				ScalableViT-S~\cite{yang2022scalablevit} & 30.0 & {45.0} & 44.9 \\
				FaViT-B2~\cite{qin2023factorization} & 29.0 & {45.2} & 45.0 \\
				PVTv2-B2~\cite{wang2022pvt} & 29.1 & 45.8 & 45.2 \\
				CrossFormer-S~\cite{wang2023crossformer++} & 34.0 & 61.0 & 46.0 \\
				\rowcolor{mygray}
				FViT-Small & {27.6} & 46.2 & {46.1} \\
				\midrule
				\midrule
				ResNet-101~\cite{he2016deep} & 47.5 & 65.1 & 38.8 \\
				PVT-M~\cite{wang2021pyramid} & 48.0 & {61.0} & 41.6 \\
				Swin-S~\cite{liu2021swin} & 53.0 & {70.0} & 45.2 \\
				LITv2-M~\cite{pan2022fast} & 52.0 & {63.0} & 45.7 \\
				FaViT-B3~\cite{qin2023factorization} & 52.0 & {66.7} & 47.2 \\
				PVTv2-B3~\cite{wang2022pvt} & 49.0 & 62.4 & 47.3 \\
				CrossFormer-B~\cite{wang2023crossformer++} & 56.0 & 91.0 & 47.7 \\
				\rowcolor{mygray}
				FViT-Base & {45.3} & 62.7 & {47.9}\\
				\bottomrule[0.125em]
		\end{tabular}}
	\end{center}
	\label{tab:ADE20K}
\end{table} 

\begin{figure}[bp]
	\centering\small\resizebox{1.0\linewidth}{!}{
		\begin{tabular}{c}
			\hspace{-3mm}\includegraphics[width=1.0\textwidth]{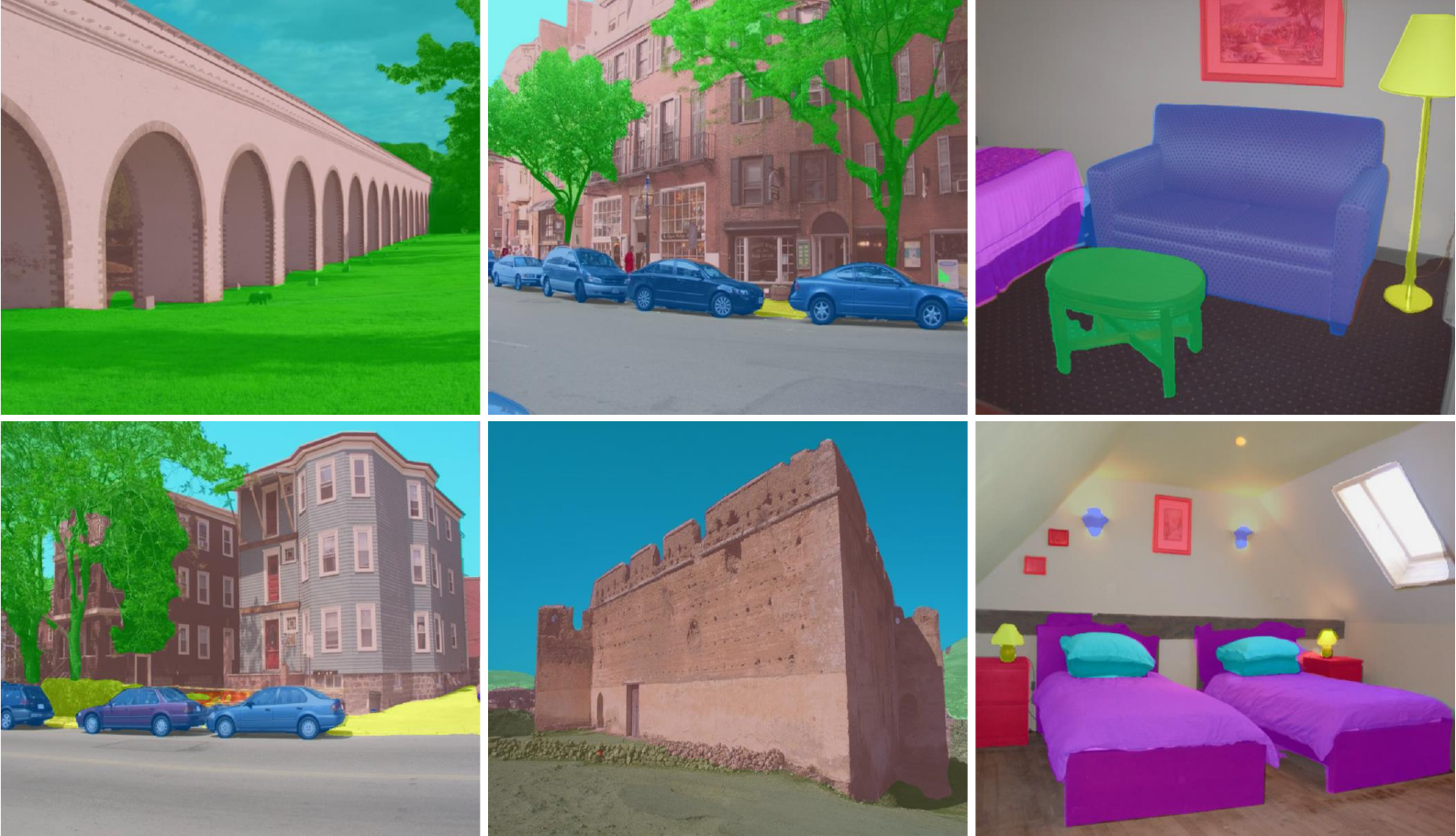}
	\end{tabular}}
	\caption{Qualitative results of FViTs for semantic segmentation on the ADE20K validation set.}
	\label{fig:ADE20K}
\end{figure}

\noindent\textbf{Results.} Table~\ref{tab:ADE20K} shows the performance comparison of FViTs with other backbone networks~\cite{liu2021swin, wang2022pvt, wang2023crossformer++} for semantic segmentation on the ADE20K~\cite{zhou2017scene} dataset. In this experiment, the model parameters, computational complexity (FLOPs), and mean Intersection over Union (mIoU) are used as evaluation metrics. Specifically, FViT-Small and FViT-Base achieves mIoU of 46.1\% and 47.9\%, respectively. Under similar model parameters and FLOPs, the FViT-Small and FViT-Base outperform the Swin~\cite{liu2021swin} and FaViT~\cite{qin2023factorization} by at least 2.7\% and 0.7\% mIoU, respectively. In addition, a more comprehensive comparison is conducted with LITv2~\cite{pan2022fast}, PVTv2~\cite{wang2022pvt}, and CrossFormer~\cite{wang2023crossformer++}. The numerical results indicate that the FViTs demonstrate a significant competitive advantage over these networks in dense prediction tasks. Furthermore, Fig.~\ref{fig:ADE20K} shows the prediction results of FViTs on the ADE20K validation set. Qualitative results show that FViTs demonstrate excellent performance in the semantic segmentation task and can effectively segment the objects of interest in complex scenarios. This further demonstrates that although FViTs do not utilize self-attention to model target features, it can still achieve satisfactory results in dense prediction tasks.

\subsection{Ablation Studies}

\noindent\textbf{Settings.} The LGF and DPFFN are the main contributions of this work and are the basic components to constitute FViTs. They enable FViTs to achieve excellent performance in image classification, object detection, and semantic segmentation tasks, while obtaining a good trade-off between computational efficiency and accuracy. In this section, the ablation experiments are conducted on ImageNet-1K dataset for validating the effectiveness of LGF and DPFFN. The same experimental setup as in Section 4.1 is followed, and FViT-Small and FViT-Base are selected as baselines.

\noindent\textbf{Results.} Table~\ref{tab:ablation} shows the ablation experimental results of LGF and DPFFN on the ImageNet-1K dataset. The numerical results show that the performance of models is improved by replacing the FFN with the DPFFN. Specifically, the performance of FViT-Small and FViT-Base improves by 0.7\% and 0.9\%, respectively, when using DPFFN instead of FFN. These results indicate that the DPFFN has better capability than FFN in modeling feature representations, especially when handling complex patterns of information interaction. In addition, the classification accuracies of FViT-Small and FViT-Base are 81.6\% and 82.5\% when using the combination of LGF and FFN, respectively. Although these results are slightly lower than those with DPFFN, they further demonstrate that LGF can serve as an effective alternative to the self-attention. Since LGF avoids the quadratic computational complexity of self-attention, it allows FViTs to run with lower computational cost while maintaining performance. This is important for efficiency optimization in practical applications, especially in scenarios with limited hardware resources or high inference speed requirements.

\begin{table}[t]
	\caption{Ablation comparison experiments of LGF and DPFFN on ImageNet-1K.}
	\begin{center}
		\resizebox{1.0\linewidth}{!}{
			\begin{tabular}{l|l|>{\centering\arraybackslash}m{1.5cm}|>{\centering\arraybackslash}m{1.5cm}|>{\centering\arraybackslash}m{1.5cm}}
				\toprule[0.125em]
				{Backbone} & {Method} & FLOPs (G) & Params (M) & Top-1 Acc (\%) \\
				\midrule
				
				\multirow{2}{*}{FViT-Small} 
				& +FFN & 3.90 & 23.28 & 81.6 \\
				& +DPFFN & 3.98 & 23.59 & 82.3 \\
				\midrule
				\multirow{2}{*}{FViT-Base} & +FFN & 7.08 & 42.62 & 82.5 \\
				& +DPFFN & 7.21 & 43.15 & 83.4 \\
				\bottomrule[0.125em]
			\end{tabular}
		}
	\end{center}
	\label{tab:ablation}
\end{table}

\section{Conclusion}

In this paper, a unified and efficient pyramid backbone network family called Focal Vision Transformers (FViTs) is proposed. The FViTs revisit the potential benefits of combining vision transformers with Gabor filters and propose a learnable Gabor filter (LGF) using convolution. The LGF is employed as an alternative to self-attention, thereby encouraging models to focus on feature representations of objects at different scales and orientations. It effectively addresses several key challenges of vision transformers, including quadratic computational complexity, high memory cost, and lack of sensitivity to local features. In addition, inspired by the biological visual system, a Dual Path Feed forward Network (DPFFN) is designed, which enables hierarchical and parallel processing of visual information, thereby expanding the receptive field within each network layer and enhancing the ability of the network to represent multiscale features at more complex levels. Both LGF and DPFFN are user-friendly, scalable, and compatible with various macro-architectures and micro-designs. Experimental results show that FViTs have significant advantages in both computational efficiency and generalization. They demonstrate good performance in image classification, object detection and semantic segmentation tasks.

\section*{Acknowledgments}

This work was partially supported by the National Natural Science Foundation of China under Grants Nos.62473209 and 62073177.

\bibliographystyle{cas-model2-names}
\bibliography{cas-refs}

\end{document}